\documentclass[10pt,twocolumn,letterpaper]{article}
\usepackage{cvpr}              % To produce the CAMERA-READY versio
\usepackage{lineno}
\usepackage{colortbl}
\usepackage{amsmath,amssymb,amsfonts}
\usepackage{graphicx}
\usepackage{float}

\usepackage{graphicx}
\usepackage{listings,newtxtt}
\usepackage{booktabs}
\usepackage{algorithm}

\usepackage{times}
\usepackage{epsfig}
\usepackage{graphicx}
\usepackage{upgreek}
\usepackage{booktabs}
\usepackage{diagbox}
\usepackage{multirow}
\usepackage{mathtools,amssymb}

\makeatletter
\@namedef{ver@everyshi.sty}{}
\makeatother
\usepackage{tikz}
\usepackage{pgfplots}
\usepackage{caption}
\usepackage{graphicx}
\usepackage{pifont}% http://ctan.org/pkg/pifont
\newcommand{\cmark}{\ding{51}}%
\usepackage{graphicx}
\usepackage{amsmath}
\usepackage{listings,newtxtt}
\usepackage{booktabs}
\usepackage{algorithm}
\newlength\savewidth

\makeatletter\renewcommand\paragraph{\@startsection{paragraph}{4}{\z@}
  {.5em \@plus1ex \@minus.2ex}{-.5em}{\normalfont\normalsize\bfseries}}\makeatother

\usepackage{algorithm}
\usepackage{algpseudocode}

\usepackage[pagebackref=true,breaklinks=true,colorlinks,bookmarks=false]{hyperref}

\usepackage[capitalize]{cleveref}
\crefname{section}{Sec.}{Secs.}
\Crefname{section}{Section}{Sections}
\Crefname{table}{Table}{Tables}
\crefname{table}{Tab.}{Tabs.}

\usepackage{xcolor} 

\def\cvprPaperID{5616} % *** Enter the CVPR Paper ID here
\def\confName{CVPR}
\def\confYear{2022}

\begin{document}

% \title{Learning Semi-supervised Semantic Segmentation by Cross-head Co-training}
\title{UCC: Uncertainty guided Cross-head Co-training for 
Semi-Supervised Semantic Segmentation}
\author{
Jiashuo Fan$^1$ \quad
Bin Gao$^2$ \quad
Huan Jin $^2$ \quad
Lihui Jiang $^2$$\thanks{Corresponding Author: jianglihui1@huawei.com}$ \\[1.2mm]
$^1$Tsinghua-Berkeley Shenzhen Institute, Tsinghua University ~~\quad
$^2$Huawei Noah’s Ark Lab
}
\maketitle
%%%%%%%%% ABSTRACT
\begin{abstract}
Deep neural networks (DNNs) have witnessed great successes in semantic segmentation, which requires a large number of labeled data for training. We present a novel learning framework called Uncertainty guided Cross-head Co-training (UCC) for semi-supervised semantic segmentation. Our framework introduces weak and strong augmentations within a shared encoder to achieve co-training, which naturally combines the benefits of consistency and self-training. Every segmentation head interacts with its peers and, the weak augmentation result is used for supervising the strong. The consistency training samples’ diversity can be boosted by Dynamic Cross-Set Copy-Paste (DCSCP), which also alleviates the distribution mismatch and class imbalance problems. Moreover, our proposed Uncertainty Guided Re-weight Module (UGRM) enhances the self-training pseudo labels by suppressing the effect of the low-quality pseudo labels from its peer via modeling uncertainty. Extensive experiments on Cityscapes and PASCAL VOC 2012 demonstrate the effectiveness of our UCC. Our approach significantly outperforms other state-of-the-art semi-supervised semantic segmentation methods. It achieves 77.17$\%$, 76.49$\%$ mIoU on Cityscapes and PASCAL VOC 2012 datasets respectively under 1/16 protocols, which are +10.1$\%$, +7.91$\%$ better than the supervised baseline.
\end{abstract}

\section{Introduction}
Image semantic segmentation is an important and hot topic in the computer vision field, which can be applied to autonomous driving\cite{qu2021focus}, medical image processing and smart city. In the past few years, semantic segmentation methods based on deep neural networks (DNNs) have made tremendous progress, such as\cite{long2015fully,chen2018encoderdecoder,chen2017deeplab,wang2020deep}. However, most of these methods involve pixel-level manual labeling, which is quite expensive and time-consuming.

To effectively utilize unlabeled images, the consistency regularization based methods have been widely used in semi-supervised learning \cite{laine2016temporal,tarvainen2017mean,french2020semisupervised}. It promotes the network to generate similar predictions for the same unlabeled image with different augmentations by calculating the difference between outputs as the loss function. Among them, data augmentation is commonly used in consistency regularization, and \cite{vanEngelen2019ASO,yang2021survey} serve a pool of data augmentation policies by designing the search space. Additionally, FixMatch\cite{sohn2020fixmatch} shows its effectiveness by enforcing consistency constraints on predictions generated by weak and strong augmentation. Despite the success of consistency regularization, we find that the network performance tends to reach its bottleneck in the high-data regime.

Another semi-supervised learning method, self-training, could fully use massive data. It incorporates pseudo labels on the unlabeled images obtained from the segmentation model to guide its learning process and then retrains the segmentation model with both labeled and unlabeled data. However, the traditional self-training procedure has its inherent drawbacks: the noise of pseudo labels may accumulate and influence the whole training process. As an extension of self-training, co-training\cite{qiao2018deep,2010Semi} lets multiple individual learners learn from each other instead of collapsing to themselves. To fully take the merit of consistency regularization and co-training, we propose a Cross-head Co-training learning framework incorporated with weak and strong augmentation. Comparing multiple models, we could achieve co-training with minimal additional parameters by the shared encoder, which enforces constraints on different learners and avoids them from converging in the opposite direction.

Our method also benefits from the learners’ diversity. Co-training will fall into self-training without diversity, and consistency training on the same prediction will also be meaningless for the lack of diversity. The diversity inherently comes from the randomness in the strong augmentation function (the unlabeled examples for two heads are differently augmented and pseudo-labeled) and different learners’ initialization. Copy-Paste (CP) is also an alternative way to boost training samples’ diversity, and recent work \cite{2010Semi} has proved its effectiveness. However, the plain CP has its inherent drawback caused by two problems. The first is the distribution mismatch between labeled data and unlabeled data. The second is the class imbalance problem, most current semantic segmentation datasets\cite{cordts2016cityscapes,Everingham15} contain long-tailed categories. By extending Copy-Paste (CP) into Dynamic Cross-Set Copy-Paste (DCSCP), our method could not only boost consistency training samples’ diversity but also reduce the misalignment between two sets’ samples and address the class imbalance problem via preserving long-tailed samples. Meanwhile, to reduce the negative effect of noisy predictions brought by self-training, Uncertainty Guided Re-weight Module (UGRM) is applied to unsupervised loss to dynamically give more weight to reliable samples while suppressing the noisy pseudo labels in self-training.

By combining data-augmentation strategies followed by the uncertainty re-weight module, we develop Uncertainty guided Cross-head Co-training, which combines consistency and co-training naturally. Experimental results with various settings on two benchmarks, Cityscapes and PASCAL VOC 2012, show that the proposed approach achieves the state-of-the-art semi-supervised segmentation performance.

Our contributions can be summarised as follow:

$\bullet$We propose a novel framework UCC, which introduces weak and strong augmentations into the cross-head co-training framework. Through the shared module, we can further improve the generalization ability and learn a more compact feature representation from two different views.

$\bullet$ A method called DCSCP is proposed to boost consistency training samples’ diversity while simultaneously reducing distribution misalignment and addressing the class imbalance problem. Furthermore, we propose UGRM to tackle the noise of pseudo labels brought by self-training.

$\bullet$ We validate the proposed method on  Cityscapes and PASCAL VOC 2012 dataset, which significantly outperforms other state-of-the-art methods with all the labeled data ratios.

%%%%%%%%% BODY TEXT
\section{Related Work}
\label{sec:intro}
% In the supervised learning research field, it's extremely difficult, expensive, or time-consuming to obtain labeled samples. One of the significant limitations to training an excellent fully-supervised deep neural network is the lack of high-quality labels.  For this reason, Semi-supervised learning (SSL) has been a hot research topic in machine learning in the last decade, where labels are not available for all training data, the aim is to leverage labeled data and a large amount of unlabeled data. So there are lots of research about semi-supervised semantic segmentat ion.

\noindent\textbf{Semi-supervised classification.} 
Semi-supervised classification methods mainly
pay attention to consistency training by combining a standard
supervised loss and an unsupervised
consistency loss to make predictions of unlabeled samples consistent under different perturbations. For example, Temporal Ensembling\cite{laine2016temporal} expects current and past epochs predictions
to be as consistent as possible. Mean teacher\cite{tarvainen2017mean} modifies Temporal Ensembling via using the average of model
weights by Exponential Moving Average (EMA) over training steps and tends to produce a more accurate model instead of directly using output predictions. Dual Student\cite{ke2019dual} further extends the Mean
Teacher model by replacing the teacher with another student.

Other studies like S4L\cite{zhai2019s4l}
explore a self-supervised auxiliary task (e.g., predicting rotations) on unlabeled images jointly with a supervised task. MixMatch\cite{berthelot2019mixmatch} produces augmented labeled and unlabeled samples
that were generated using Mixup\cite{zhang2018mixup}. FixMatch\cite{sohn2020fixmatch} leverages consistency regularization between weakly and strongly augmented views of the same unlabeled image in a single-stage training pipeline.

\begin{figure*}[t]
\centering
\footnotesize
\includegraphics[width=1\linewidth]{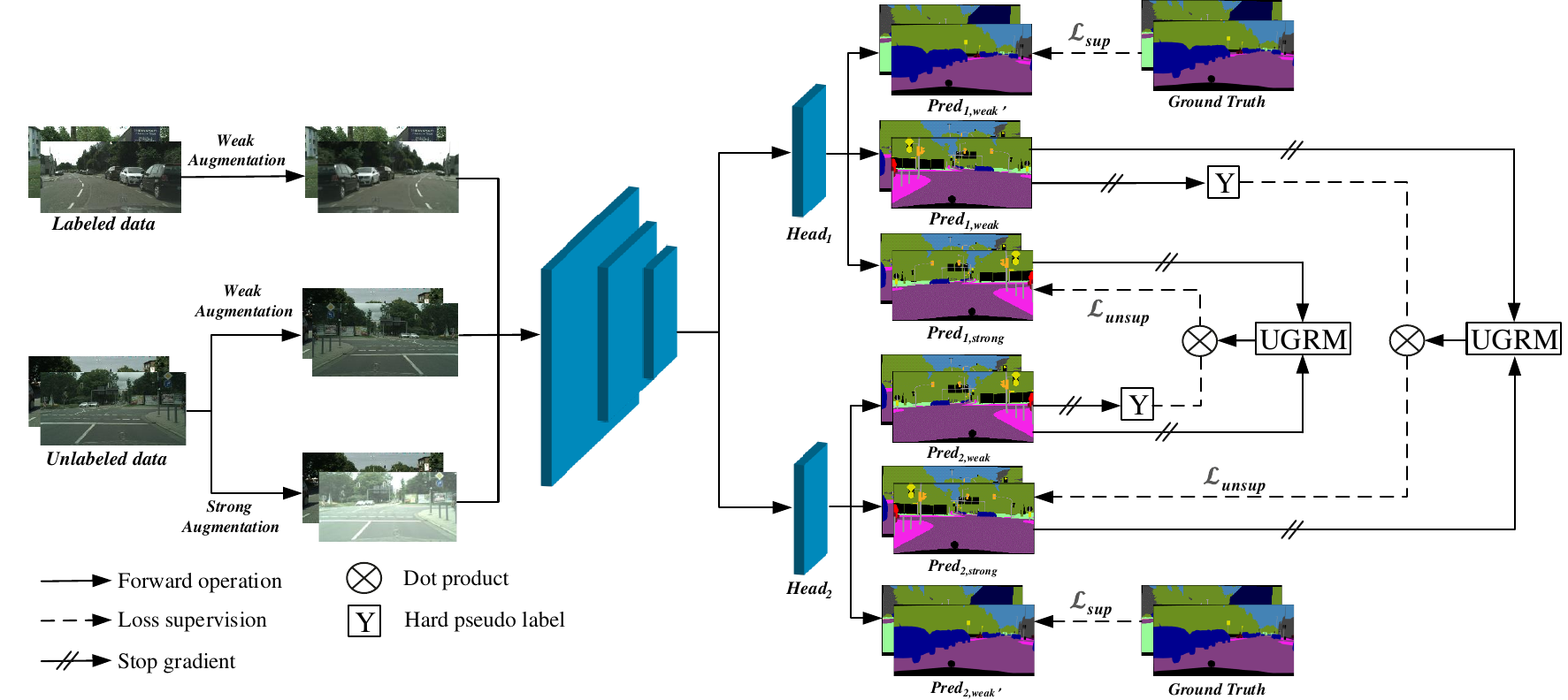}
\captionsetup{font={small}}
\caption{
Overview of our Uncertainty guided Cross-head Co-training. In our scheme, we use $head_1$ as an example, weakly augmented label data will flow through a shared module and corresponding segmentation $head_1$ to generate ${Pred}_{{1,weak}^{'}}$, then it will be supervised by Ground Truth (GT). For unlabeled data, a weakly augmented image is fed into a shared module and corresponding segmentation $head_1$ to generate ${Pred}_{{1, weak}}$. Different from ${Pred}_{{1,weak}^{'}}$,  ${Pred}_{{1,weak}}$ is for supervision signal towards strong image predictions from the other head ${Pred}_{{2,strong}}$, vice versa. Additionally, to further reduce the impact of the noisy label, UGRM is involved in the training process to tackle the noise of the pseudo label.
}
\label{fig:structure}
\vspace{-4mm}
\end{figure*}

\noindent\textbf{Semi-Supervised Semantic Segmentation.} Inspired by
the recent progress of SSL methods in the image classification domain, a few works explore semi-supervised
learning in semantic segmentation and show promising results. 
Preliminary
works \cite{mittal2019semi,hung2018adversarial} in semi-supervised semantic segmentation tend to utilize the Generative Adversarial 
Networks (GANs)\cite{creswell2018generative} as an auxiliary supervision signal for the unlabeled data in a way that the discriminator judges the
produced predictions to imitate the common structures and semantic
information of true segmentation masks. Alternative approaches like Contrastive learning also show up prominently. ReCo\cite{liu2021bootstrapping} turns to offer relevance for semantic class under contrastive framework. PC2Seg\cite{zhong2021pixel} leverages the label-space consistency
property between image augmentations and the feature space contrastive property among different pixels. C3-SemiSeg\cite{Zhou_2021_ICCV} achieves contrastive learning by computing the loss of pixel-wise features and adopts a sample strategy to tackle the noisy labels through a class-wise ema threshold.

Consistency regularization methods are also commonly used in semi-supervised semantic segmentation. CCT\cite{ouali2020semi} introduces a feature-level perturbation
and enforces consistency among the predictions of different decoders. GCT\cite{ke2020guided} performs network
perturbation by using two differently initialized segmentation models and encourages consistency
between the predictions from the two models. GuidedMix-Net\cite{tu2021guidedmix} builds upon\cite{french2020semisupervised,zhang2018mixup} and enforces the mixed
predictions and predictions of mixed inputs to be consistent
with each other.  

Self-training via pseudo labeling is a classical method originating
around a decade ago, taking the most probable class as a pseudo label and training models on unlabeled data, which is a frequently-used approach to achieve minimum
entropy. It's firstly used for classification task\cite{radosavovic2018data,cascante2020curriculum,xie2020self,yalniz2019billion}. Recently, it's been widely applied in semi-supervised semantic segmentation tasks such as \cite{feng2020dmt,zhai2019s4l,yuan2021simple}. \cite{feng2020dmt} re-weights the loss on different regions based on the disagreement
of two different initialized models. \cite{zhai2019s4l} selects and prioritizes
easy or reliable images during the re-training stage to exploit the unlabeled images.\cite{yuan2021simple} adopts heavy augmentations to achieve high performance.

Though there are many strategies aiming to utilize labeled and unlabeled data for improving the model performance\cite{ghiasi2021simple}, the labeled and unlabeled data distribution mismatch has rarely been discussed. In fact, the empirical distribution of labeled data often deviates from the true samples' distribution\cite{wang2019semisupervised}, the model performance will be significantly degraded when there exists a considerable distribution mismatch. In addition to it, long-tailed class distribution is a common issue in semi-supervised semantic segmentation. There are few researches about this problem in semi-supervised semantic segmentation\cite{wei2021crest,hu2021semi}, while \cite{wei2021crest} introduces a self-training framework for imbalanced category. \cite{hu2021semi} proposes a framework to encourage under-performing categories to be sufficiently trained by using a dynamic confidence bank.

\section{Semi-supervised learning with cross-head co-training}

In this section, we present an overview of the proposed framework UCC in Section \ref{sec:Overview}. Then we describe our weak and strong data augmentation strategy in Section \ref{sec:Weak and Strong augmentation}. Additionally, we propose a dynamic copy-paste with cross-set data mixing strategy in Section \ref{sec:Dynamic Cross-Set Copy-Paste Strategy}. Finally, an uncertainty guided loss re-weight module is further introduced in Section \ref{sec:Uncertainty Estimation}.

\subsection{Overview}
\label{sec:Overview}

Figure \ref{fig:structure} is a visualization of our cross-head architecture. Images are fed into a shared CNN backbone followed by two same segmentation heads. Compared with using separate individual models, cross-head could learn a compact feature and further improve generalization ability. For a labeled image, we compute the supervised loss $L_{sup}$ between GT and accordingly prediction of weak version. For an unlabeled image, the pseudo label is generated by weakly augmented image prediction. The pseudo label is then used to supervise the strongly-augmented image prediction from the other head. On the one hand, the pseudo label plays a role in expanding training data. On the other hand, enforcing the constraint for weak and strong image predictions could enjoy the merit of consistency training. In addition, as shown in the figure, UGRM is added after getting unsupervised loss $L_{unsup}$. UGRM encourages more reliable samples involved in our training process, while the high-uncertain sample brought by self-training will be given less weight during training. Another module, DCSCP boosts consistency training samples' diversity while simultaneously dealing with long-tailed and distribution misalignment problems. Further details are in Section \ref{sec:Dynamic Cross-Set Copy-Paste Strategy}.

Our scheme is composed of a shared backbone $f$ and two segmentation heads $g_m$ ($m\in\{1,2\}$), where the structures of the two heads are the same as each other. Here our pseudo label is generated by $p_m(y\mid x)=g_m(f(x))$, which is produced by $f$ and $g_m$. The pseudo label will then be used as the supervision signal for the other head.

Following the setting of semi-supervised semantic segmentation, we are provided with a batch of labeled examples $\mathcal{D}_l={\{(x_b,y_b);b\in (1,\dots,B_l)\}}$ and a batch of unlabeled examples $\mathcal{D}_u=\{(u_b);b\in (1,\dots,B_u)\}$ during every iteration. We preliminarily define $\ell_{ce}$ as the standard pixel-wise cross-entropy loss, $\mathcal{W}$ and $\mathcal{S}$  represent the corresponding weak and strong augmentation function applied on the image. 
Similar to previous semi-supervised methods\cite{Zhou_2021_ICCV}, for the labeled part, the supervision loss $\mathcal{L}_s$ is formulated using the standard pixel-wise
cross-entropy loss on the labeled images over the two heads:

\begin{equation*}
    \mathcal{L}_s=\frac{1}{N_l}\sum_{i=1}^{N_l}\frac{1}{WH}\sum_{j=1}^{WH}(\ell_{ce}(\boldsymbol{y}_{ij},p_{1,ij}^{\mathcal{W}})\notag\\+\ell_{ce}(\boldsymbol{y}_{ij}, p_{2,ij}^{\mathcal{W}}))
\end{equation*}

Where $p_{m,ij}^{\mathcal{W}}=g_m(f(\mathcal{W} \circ \boldsymbol{x}_{ij}))$  represents $j$-th pixel in the $i$-th weakly-augmented labeled image prediction generated by head-$m$, ${y_{ij}} \in R$ is the corresponding  ground truth of the $j$-th pixel in the $i$-th labeled (or unlabeled) image, $N_l$ is the total number of samples in the labeled training set.

\begin{equation}
    q_{m,ij}^\mathcal{W} = \mathop{\arg\max}_c p_m(y=c\mid p_{m,ij}^{\mathcal{W}}), \\
\end{equation}

\begin{equation}
    \mathcal{L}_u = \frac{1}{N_u}\sum_{i=1}^{N_u}\frac{1}{WH}\sum_{j=1}^{WH}(\ell_{ce}(\boldsymbol{q}_{1,ij}^\mathcal{W}, p_{2,ij}^{\mathcal{S}}) \notag\\
    +\ell_{ce}(\boldsymbol{q}_{2,ij}^\mathcal{W}, p_{1,ij}^{\mathcal{S}})),\label{eq:Lu}
\end{equation}

For unlabeled data, an unsupervised loss combined with consistency and self-training is applied to encourage consistent pseudo label predictions in response to one image with different perturbations. Argmax function chooses the corresponding class $c\in\{1,\dots,C\}$ with the maximal probability. $p_{m,ij}^{\mathcal{S}}=g_m(f(\mathcal{S}\circ\boldsymbol{x}_{ij}))$ denotes $j$-th pixel in the $i$-th strongly-augmented unlabeled image prediction generated by head-$m$, $N_u$ is the total number of unlabeled samples in the training set.

Finally, the whole loss is written as:
\begin{align}
\label{eq:whole-loss}
\mathcal{L} = \mathcal{L}_s
+ \lambda \mathcal{L}_{u},
\end{align}

Here we use $\lambda$ to control the balance between supervised and unsupervised loss.

\lstset{
  backgroundcolor=\color{white},
  basicstyle=\fontsize{8.5pt}{9.5pt}\fontfamily{lmtt}\selectfont,
  columns=fullflexible,
  breaklines=true,
  captionpos=b,
  commentstyle=\fontsize{9pt}{10pt}\color{codegray},
  keywordstyle=\fontsize{9pt}{10pt}\color{codegreen},
  stringstyle=\fontsize{9pt}{10pt}\color{codeblue},
  frame=tb,
  otherkeywords = {self},
}

\subsection{Weak and Strong augmentation} 
\label{sec:Weak and Strong augmentation}
% In our framework, several data augmentation techniques are proposed to facilitate our training; The following are the details.
 
% \noindent\textbf{Weak and Strong Strategy.}  
To fully enjoy the merit of consistency training, we leverage weak and strong augmentation to introduce extra information in our framework. In our experiments, weak augmentation is the combination of standard flip-and-shift, random scale and crop strategy. Specifically, we randomly
flip and scale images with a probability of $50\%$.

Our method produces pseudo labels using both consistency regularization and self-training pseudo labeling.
Specifically, the pseudo label is produced based on the weakly-augmented unlabeled image, which is then used as the supervision signal when the model is fed a strongly augmented version of the same image. Similar to RandAugment, as illustrated in Figure \ref{fig: strong aug}, we build
a pool of operations containing nine image transformations. Instead of using a fixed global magnitude during every training iteration, we randomly select transformations for each sample in a mini-batch from a pre-defined range at each training step.

% \chapter[Title]{\color{Maroon}Title}
\definecolor{codeblue}{rgb}{0, 0, 0}
\definecolor{codegray}{rgb}{0, 0, 0}
\definecolor{codegreen}{rgb}{0, 0, 0}
\begin{figure}[h]
\begin{lstlisting}[language=python]
Strong_Transforms = ['channel_shuffle','color_jitter', 'invert_channel','coarse_dropout','jpeg_compression','solarize','salt_pepper','noise_gau','equalize']

Weak_Transforms = ['random_scale', 'random_flip', 'random_crop', 'Normalization']

# define strong augmentation func
def Strong_augmentation(N):   
""" Generate a set of distortions.
    Args: 
    N: Num. of augment. transform. to apply sequentially
"""
    Strong_ops = np.random.choice(Strong_Transforms, N)
    return Strong_ops + Weak_Transforms
\end{lstlisting}
\caption{Python code for Strong Data Augmentation %
in Numpy.}
\label{fig: strong aug}
\end{figure}

\subsection{Dynamic Cross-Set Copy-Paste Strategy}
\label{sec:Dynamic Cross-Set Copy-Paste Strategy}

Copy-Paste\cite{ghiasi2021simple} is a successful method that copies objects from one image to another, copying the specific pixels belonging to the target object instead of the rectangle mask. Copy-Paste initially aims at establishing a data-efficient model that can handle rare object categories, which offers an opportunity
to better leverage labeled data by creating a variety of new,
highly perturbed samples and then using these for training. In addition, recent work \cite{hu2021semi} adopts an adaptive way to utilize copy-paste strategy. It proposes a framework to encourage under-performing categories to be sufficiently trained using a dynamic confidence bank.

However, our target scope is significantly different. Though \cite{hu2021semi} enlarges intra-level labeled samples' diversity, it ignores the labeled and unlabeled data distribution mismatch problem. The network performance will be degraded if there exists a large gap between labeled and unlabeled data. Recent works like \cite{xu2019adversarial,wang2019semisupervised} show the effectiveness of data mix-up methods. Thus we propose our DCSCP to address labeled and unlabeled data distribution misalignment and long-tailed problems via extending copy-paste strategy. The key idea here is that we form new, augmented samples by copying all pixels belonging to the specific category and paste them both on the labeled and unlabeled image, and the pixels are sampled from an estimated labeled data confidence distribution. The corresponding mixing procedure is:

\begin{gather}
     %\mathcal
     {{x}}_{copy\_paste} = M\odot x^a+(1-M)\odot x^b, 
\end{gather}

Given two images ${ {x}^a \in {D}_a}$, $ {x}^{b} \in {D}_b$, where $\mathcal {D}_b \subset D$. We extend $D$ into ${D}_l\cup{D}_u$ instead of ${D}_l$ so that unlabeled data could also share the spirit of copy-paste by addressing class imbalance problem. Here ${\mathcal {D}_a} \subset {D}_l$ stays the same and M denotes the copy-paste semantic mask of pixels belonging to one specific category. 

Specifically, for each forward pass, we compute the  average pixel-wise confidence distribution for the c-th
class as $\mathcal{\hat{\sigma}}_{t,c}$. Then the class-wise confidence distribution is
updated through an exponential moving average way:

\begin{gather}
     \mathcal{\hat{\sigma}}_{t,c} = \alpha\mathcal{\hat{\sigma}}_{t-1,c}+(1- \alpha){\sigma}_{t,c}, \label{eq:gamma}
\end{gather}

Where $\mathcal{\alpha}$ denotes the ema ratio, ${\hat{\sigma}}_{t,c}$ denotes the average confidence distribution for the c-th class at t-step, which is
smoother by the past threshold information. Then the confidence distribution is used to the category selecting procedure.

\renewcommand{\arraystretch}{1.5}
\begin{table*}[]
\centering
\setlength{\tabcolsep}{7.5pt}
\captionsetup{font={small}}
\caption{Comparison with state-of-the-arts
on the Cityscapes val set under
different partition protocols.
{\color{black}All the methods are based on DeepLabv$3$+.}
}
\vspace{-2mm}
\centering
\label{tab:sota-City-subset}
\footnotesize
\begin{tabular}{l|c|c|c|c|c|c|c|c}
\hline 
\multirow{2}{*}{Method} & \multicolumn{4}{c|}{ResNet-$50$} & \multicolumn{4}{c}{ResNet-$101$} \\\cline{2-9}
& $1/16$ ($186$) & $1/8$ ($372$) & $1/4$ ($744$) & $1/2$ ($1488$) & $1/16$ ($186$) & $1/8$ ($372$) & $1/4$ ($744$) & $1/2$ ($1488$) \\
\hline
Baseline~ &  $65.28$ & $71.33$ & $73.78$ & $76.04$ & $67.16$ & $72.30$ & $74.60$ & $76.7$\\
\hline
MT*\cite{tarvainen2017mean} &  $66.14$ & $72.03$ & $74.47$ & $77.43$ & $68.08$ & $73.71$ & $76.53$ & $78.59$ \\
CCT*~\cite{ouali2020semi} & $66.35$ & $72.46$ & $75.68$ & $76.78$ & $69.64$ & $74.48$ & $76.35$ & $78.29$ \\
GCT*~\cite{ke2020guided} & $65.81$ & $71.33$ & $75.30$ & $77.09$ & $66.90$ & $72.96$ & $76.45$ & $78.58$\\ 
CPS~\cite{chen2021semi} & ${74.47}$ & $ {76.61} $ & ${77.83} $ & $ {78.77} $ & ${74.72}$ & $ {77.62} $ & ${79.21} $ & $ {80.21} $ \\ 
\hline
Ours   & $\mathbf{76.02}$ & $ \mathbf{77.60} $ & $\mathbf{78.28} $ & $ \mathbf{79.54} $ & $\mathbf{77.17}$ & $ \mathbf{78.71} $ & $\mathbf{79.59} $ & $\mathbf{80.57} $ \\ 
\hline
\end{tabular}
\vspace{-2mm}
% \vspace{-0.3cm}
\end{table*}

\renewcommand{\arraystretch}{1.5}
\begin{table*}[]
\centering\setlength{\tabcolsep}{6.5pt}
\captionsetup{font={small}}
\caption{Comparison with state-of-the-arts
on the PASCAL VOC $2012$ val set
under different partition protocols.
{\color{black}All the methods are based on DeepLabv$3$+.}
}
\vspace{-2mm}
\label{tab:sota-VOC-subset}
\footnotesize
\begin{tabular}{l|c|c|c|c|c|c|c|c}
\hline 
\multirow{2}{*}{Method} & \multicolumn{4}{c|}{ResNet-$50$} & \multicolumn{4}{c}{ResNet-$101$} \\\cline{2-9}
& $1/16$ ($662$) & $1/8$ ($1323$) & $1/4$ ($2646$) & $1/2$ ($5291$) & $1/16$ ($662$) & $1/8$ ($1323$) & $1/4$ ($2646$) & $1/2$ ($5291$) \\
\hline
{\color{black}Baseline}  & $65.28$ & $68.09$ & $72.35$ & $74.00$ & $68.58$ & $72.56$ & $75.05$ & $76.03$ \\
\hline
{\color{black}MT*}~\cite{tarvainen2017mean}   & $66.77$ & $70.78$ & $73.22$ & $75.41$ & $70.59$ & $73.20$ & $76.62$ & $77.61$ \\
CCT*~\cite{ouali2020semi} &  $65.22$ & $70.87$ & $73.43$ & $74.75$ & $67.94$ & $73.00$ & $76.17$ & $77.56$ \\
{\color{black} GCT*}~\cite{ke2020guided} & $64.05$ & $70.47$ & $73.45$ & $75.20$ & $69.77$ & $73.30$ & $75.25$ & $77.14$\\
{\color{black} CPS}~\cite{chen2021semi} & $71.98$ & $ 73.67$ & $74.9$ & $76.15$ & $74.48$ & $76.44$ & $77.68$ & $78.64$ \\ \hline
Ours & $\mathbf{74.05}$ & $ \mathbf{74.81} $ & $\mathbf{76.38} $ & $\mathbf{76.53} $ & $\bf{76.49}$ & $\bf{77.06}$ & $\bf{79.07}$ & $\bf{79.54}$ \\ \hline
\end{tabular}
\vspace{-2mm}
\end{table*}

\subsection{Uncertainty Estimation}
\label{sec:Uncertainty Estimation}
% Disagreement based strategy is often followed by assumptions that Multiple decoders should reach an agreement on
% unlabeled data predictions[15].
% We can divide the disagreement method into two types when maximizing their agreement on unlabeled data.
Because of the noise of pseudo labels, even a tiny error tends to accumulate to the degree that will degrade the model performance by a large margin. To tackle the noisy labels. Previous work like\cite{feng2020dmt} adopts a hard-voting method based on the majority wins. However, arbitrarily using consensus results generated by the majority wins may lead to error accumulation. \cite{french2020semisupervised,ouali2020semi} show their effectiveness by using a fixed threshold to filter noisy pseudo labels. Nevertheless, the unchanged and fixed threshold ignores some useful pseudo labels under the threshold. 

Though these methods show their effectiveness in dealing with the noisy labels to some extent, they are still hindered by their inherent weakness. Hence we propose UGRM to address noisy problems via modeling uncertainty based on the soft-voting paradigm. To be specific, we consider how certain each learner is and take the target class as ground truth when the probability value from its peer is higher. Therefore, our method could not only effectively mitigate the negative effect of pseudo labels but also alleviate the error accumulation problem. We first re-weight pixel-level loss by:

\begin{gather}
     %\mathcal 
     w_{m,ij} = \max_{c\in \{1,\dots,C\}}p_{m,ij}^c, \label{eq:gamma}\\
     u_{m,ij}^1 = w_{m,ij}, \label{eq:u2}
\end{gather}

Where $w_{m,ij}$ denotes the maximal probability within the class $c\in \{1,\dots,C\}$. $u_{m,ij}^1$  is computed by Eq \eqref{eq:u2}, it's our first re-weight factor to control the contribution of every-pixel.

\begin{gather}
     %\mathcal 
     u_{1,ij}^2 = 1_{ij},w_{2,ij} >w_{1,ij}, \label{eq:u11}\\
     u_{2,ij}^2 = 1_{ij},w_{1,ij} >w_{2,ij}, \label{eq:u12}
\end{gather}

In Eq \eqref{eq:u11}, Eq \eqref{eq:u12},
%      u_{m,ij} = u_{m,ij}^1 * u_{m,ij}^2
$1_{ij}=1$ represents that if the current head's prediction confidence of $j$-th pixel in the $i$-th image is higher compared with others, then it is equal to 1; otherwise, $1_{ij}$ will be 0. It is sensible that the confidence of pseudo label should be used as a supervision signal when it is reliable. Otherwise, it should be discarded during the training process. Combining $u_{m,ij}^1,u_{m,ij}^2$ and merging it into our unsupervised loss, we can derive Eq \eqref{eq:final} and rewrite Eq \eqref{eq:Lu}  as follows, which can not only alleviate the impact of noise through weight adjustment but also feed the chosen reliable samples into training decided by two heads.

\begin{gather}
    u_{m,ij} = u_{m,ij}^1 *u_{m,ij}^2,\label{eq:final}\\
    \mathcal{L}_u = \frac{1}{N_u}\sum_{i=1}^{N_u}\frac{1}{WH}(\frac{1}{\sum_{j=1}^{WH}u_{1,ij}}\sum_{j=1}^{WH}u_{1,ij}\ell_{ce}(\boldsymbol{q}_{1,ij}^\mathcal{W}, p_{2,ij}^{\mathcal{S}}) \notag\\
    +\frac{1}{\sum_{j=1}^{WH}u_{2,ij}}\sum_{j=1}^{WH}u_{2,ij}\ell_{ce}(\boldsymbol{q}_{2,ij}^\mathcal{W}, p_{1,ij}^{\mathcal{S}}))\notag
\end{gather}

 %%%%%%%%% BODY TEXT
\section{Experiments}
\subsection{Experimental Setup}
       
\noindent\textbf{Datasets.} Our main experiments and ablation studies are based on the Cityscapes dataset \cite{cordts2016cityscapes}, which contains 5K fine-annotated images. And the images are divided into training, validation, testing sets which contain 2975, 500, and 1525 images respectively. Cityscapes defines 19 semantic classes of urban scenes. Additionally, we test the proposed method on the PASCAL VOC 2012 dataset (VOC12) \cite{Everingham15}, which consists of 20 semantic classes and one background class. The standard VOC12 includes 1464 training images, 1449 validation images, and 1456 test images. Following common practice, we use the augmented set \cite{hariharan2011semantic} which contains 10, 582 images as the training set. 

We follow the partition protocols of \cite{chen2021semi} and divide the whole training set into two groups via randomly sub-sampling 1/2, 1/4, 1/8 and
1/16 of the whole set as the labeled set and regard the remaining images as the unlabeled set.

\noindent\textbf{Evaluation.} Our performance evaluation is based on single-scale testing and the mean of intersection over union (mIoU). We report the results of the Cityscapes val set and the PASCAL VOC 2012 val set compared with state-of-the-art methods. We compare our results with recent reports in a fair manner. We separately use ResNet-$50$, ResNet-$101$ as our backbone networks. The pretrained model is initialized with the supervised data. In addition, we use DeepLabv3+\cite{chen2017rethinking} as a segmentation head. We use mini-batch SGD with momentum for the Cityscapes dataset to train our model with Sync-BN. In particular, we adopt a learning policy with an initial learning rate of 0.004 which is then multiplied by $(1-\frac{iter}{max\_iter})^{0.9}$, weight decay of 0.0005, and momentum of 0.9.
We set the crop size as 800 × 800, batch size as 8. For PASCAL VOC 2012 dataset, we set the initial learning rate as 0.0005, weight decay as 0.0005, crop size as 512 × 512, batch size as 8. We use random horizontal flip, random scale and crop as our default data augmentation, and OHEM loss is used on Cityscapes and VOC12.

\subsection{Comparison to State-of-the-Art Methods}
In this section, we extensively compare our frameworks with previous methods across
various datasets and settings. The other compared state-of-art results are from \cite{chen2021semi} and marked as *. 

\noindent\textbf{Cityscapes.} In Table ~\ref{tab:sota-City-subset}, we present our results of mean Intersection over Union (mIoU) on the Cityscapes validation
dataset under different proportions of labeled samples. We
also show the corresponding baseline at the top of the table, which denotes the purely supervised learning results trained by the same labeled data. Note that all the
methods use the DeepLab V3+ for a fair comparison.

As we can see, our method is consistently superior to the supervised baseline on Cityscapes.
To be specific, compared with the baseline,  the improvements of our method are
$10.74\%$, $6.27\%$, $4.50\%$, and $3.50\%$
with ResNet-$50$ under $1/16$, $1/8$, $1/4$, and $1/2$ partition protocols separately, $10.01\%$, $6.41\%$, $4.99\%$, and $3.87\%$ with ResNet-$101$ under $1/16$, $1/8$, $1/4$, and $1/2$ partition protocols separately.

When the ratio of the labeled data becomes smaller (e.g.,
1/8,1/16), it is observed that our method behaves significant improvement in performance. Significantly, under extremely few data settings, especially under 1/16 partition, the gain from our method is $10.01\%$ over baseline, which surpasses the previous state-of-art method \cite{tarvainen2017mean} in a large margin $+2.45\%$ while using ResNet-$101$ as the backbone, and by $+1.09\%$ under 1/8 partition. In summary, our proposed method shows significant improvements
in various situations. Our method could handle massive unlabeled data and still keep good performance by combining consistency regularisation
and self-training during the training process.

\noindent\textbf{PASCOL VOC 2012.}
To further prove the generalization ability of
our method, we also conduct experiments on PASCAL VOC 2012 val dataset. As we can see from Table ~\ref{tab:sota-VOC-subset}, our method consistently beats supervised baseline over a large margin, the improvements are $8.77\%$, $6.72\%$, $4.03\%$, and $2.53\%$
with ResNet-$50$ under $1/16$, $1/8$, $1/4$, and $1/2$ partition protocols separately, $7.91\%$, $4.50\%$, $4.02\%$, and $3.51\%$ with ResNet-$101$ under $1/16$, $1/8$, $1/4$, and $1/2$ partition protocols separately. Additionally, our method is superior to all the other state-of-art methods over different settings. To be specific, it outperforms previous state-of-art \cite{chen2021semi} by $2.01\%$ and $1.39\%$ under the 1/16 and 1/4 partitions.

\subsection{Ablation Studies}
In this subsection, we will discuss every component's contribution to our framework. If not mentioned by purpose, all methods are based on DeepLabv3+ with ResNet50 under 1/8 partition protocols\cite{chen2021semi}.

\noindent\textbf{Effectiveness of the different components.} To further understand the advantages brought by different components. We conduct ablation studies step by step and progressively examine every component's effectiveness. Table ~\ref{tab:abla-of-losses} reports the results. By training plain cross-head framework without strategy described in section \ref{sec:Weak and Strong augmentation}, \ref{sec:Dynamic Cross-Set Copy-Paste Strategy}, \ref{sec:Uncertainty Estimation}, we could achieve $72.23\%$ mIoU. Besides this, the result from the plain framework could be further improved $2.07\%$ by WS, which can be attributed to taking the merit of consistency training by enforcing the constraint for weakly and strongly images' predictions. On top of WS, by merging DCSCP into our framework, it further obtains improvements of $1.40\%$ by alleviating class balance and distribution misalignment problems. Additionally, UGRM further boosts performance to $77.60\%$ through re-considering the importance of every pixel and the different information brought by two segmentation heads, which shows the effectiveness of our uncertainty estimation method.
\renewcommand{\arraystretch}{1.6}
\begin{table}[]
\centering\setlength{\tabcolsep}{15pt}
% \centering
\captionsetup{font={small}}
\caption{Ablation study on the effectiveness of different components, including
WS (Weak and Strong Strategy), UGRM (Uncertainty-Guided Re-weight Module), DCSCP (Dynamic Cross-Set Copy-Paste).
}

\vspace{-2mm}
\label{tab:abla-of-losses}
\footnotesize
\begin{tabular}{c|c|c|c}
\hline
WS  & DCSCP & UGRM &  mIoU\\ \hline
  &   &  &  $72.23$\\ 
 \cmark &   &  &   $74.3$\ \\
%   & \cmark &  &  &   $72.84$ \\
%   & &\cmark  &  &   $73.85$ \\ 
%   &  &  &  \cmark &  $73.65$ \\
  \cmark  &\cmark &  & $75.7$ \\
  \cmark  & \cmark & \cmark & $\bf{77.60}$  \\ \hline
\end{tabular}
\end{table}

\renewcommand{\arraystretch}{1.6}
\begin{table}[]
\centering\setlength{\tabcolsep}{15pt}
\captionsetup{font={small}}
\caption{Effectiveness of the weak-strong augmentation strategy. Performance analysis over different augmentation strategies.
${\mathcal{S}}$: Strong augmentation, ${\mathcal{W}}$: Weak augmentation.
\vspace{-2mm}
}
\footnotesize
\begin{tabular}{l|c}
\hline
% \multirow{2}{*}{Method} & \multicolumn{1}{c}{$\#$(labeled samples)} \\ \cline{2-2}
% & 1/8  \\
Method~  & mIoU  \\
\hline
N.A.~  & $71.32$  \\
${\mathcal{S}}$ &  $70.02$  \\
${\mathcal{W}}$~ &  $73.01$   \\
${\mathcal{W}+\mathcal{S}}$~ & $\bf{74.2}$ \\
\hline
\end{tabular}
\label{tab:voc_sota_ours_unlabeled}
% \vspace{-0.3cm}
\end{table}

\begin{figure*}[t]
\centering
\footnotesize
% Qualitative.pdf quantitive.pdf
\includegraphics[width=1\linewidth]{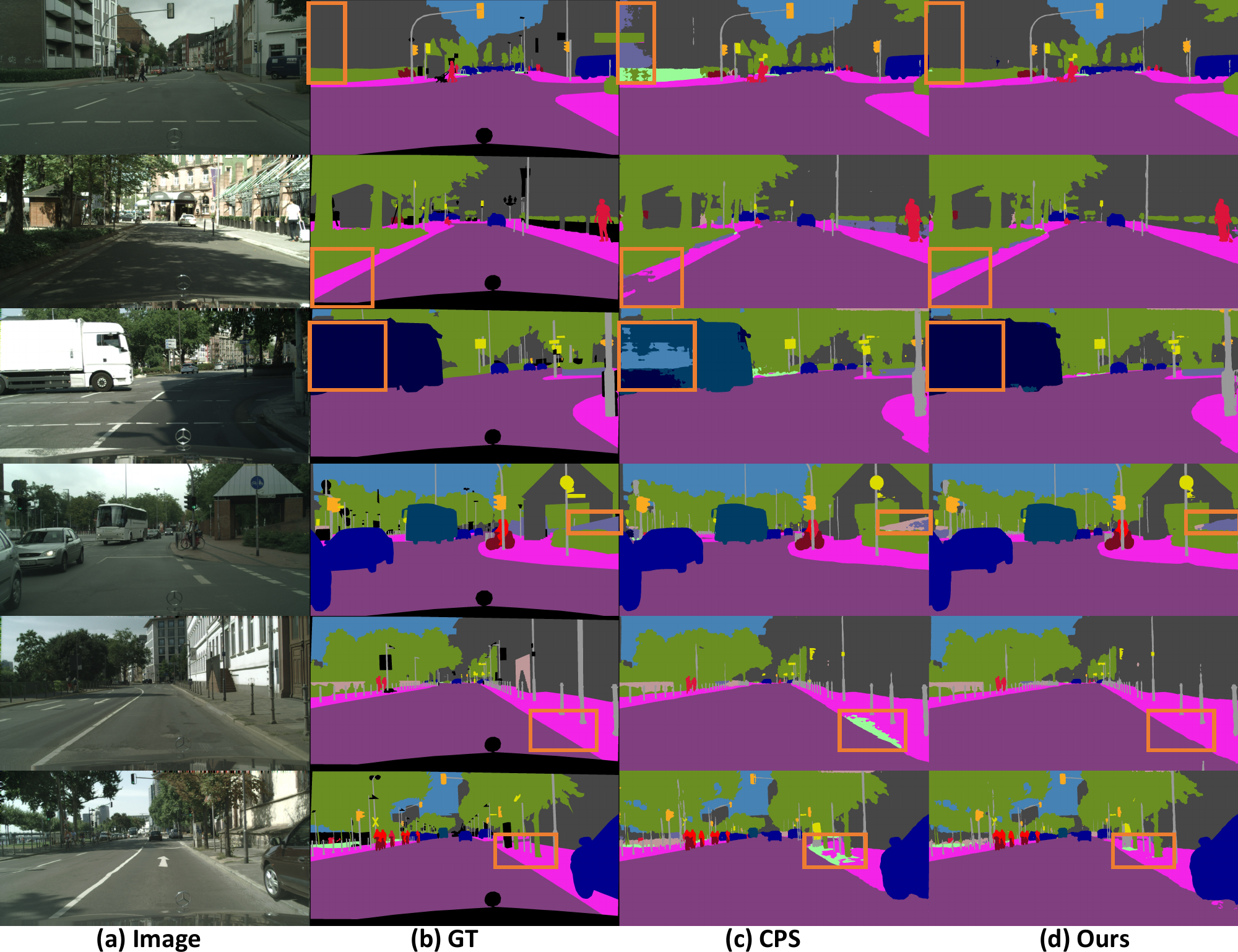}
\captionsetup{font={small}}
\caption{Qualitative results on the Cityscapes val set.(a) and (b) are corresponding to target images and Ground Truth(GT), (c) represents the results of CPS\cite{chen2021semi}, (d) is our method results. Orange rectangles highlight the difference among of them.
}
\label{fig:quantitive}
\vspace{-4mm}
\end{figure*}

\noindent\textbf{Effectiveness of the weak-strong augmentation strategy.} By utilizing the weak-strong strategy, we can introduce more information to consistency (It is worth noting that the strong transformation function generates different augmented images each time).
We conduct experiments of adding the different intensities of augmentation.
As can be seen in Table ~\ref{tab:voc_sota_ours_unlabeled}, directly applying strongly augmented prediction as supervision signal
leads to the performance drop. It may be caused by the significant
increase of wrong predictions from the other head, which misleads the network's optimization direction. To get more sensible and accurate pseudo labels, a natural idea is to generate pseudo labels by weakly-augmented versions of unlabeled images compared with strongly-augmented images
. As shown in Table ~\ref{tab:voc_sota_ours_unlabeled}. After replacing supervision signal with weakly-augmented versions to match the weakly-augmented predictions, the performance is improved by $1.69\%$. Finally, the result could be further improved by $2.88\%$ when conducting weak and strong augmentation on both heads, which shows the effectiveness of our strategy.

\noindent\textbf{Effectiveness of the DCSCP.}
We ablate each component of DCSCP step by step. Table~\ref{tab:DCSCP} shows that directly using intra-level copy-paste could bring $1.00\%$ boosts by forming new, perturbated samples. On top of CP, further extending our method into DCP yields $0.27\%$ gains, which may be attributed to sampling the target category from the estimated class distribution and encouraging rare categories to be sufficiently trained. In addition, extending DCP across both labeled and unlabeled data brings $0.62\%$ extra improvements, this can be attributed to addressing the labeled and unlabeled data distribution mismatch problem. Finally, our method improves the performance by $1.78\%$ through combining DCP and CSCP, which illustrates the proposed DCSCP is a more powerful tool for semi-supervised semantic segmentation.

\renewcommand{\arraystretch}{1.4}
\begin{table}[]
\centering\setlength{\tabcolsep}{32pt}
\captionsetup{font={small}}
\caption{Effectiveness of the DCSCP. Performance analysis over DCSCP. DCP: Dynamic Copy-Paste, CSCP: Cross-Set Copy-Paste, DCSCP: Dynamic Cross-Set Copy-Paste.
\vspace{-2mm}
}
\footnotesize
\begin{tabular}{l|ccc}
% \hline
% \multirow{2}{*}{Method} & \multicolumn{1}{c}{$\#$(labeled samples)} \\
% \cline{2-2}
% & 1/8   \\
\hline
% \multirow{2}{*}{Method} & \multicolumn{1}{c}{$\#$(labeled samples)} \\ \cline{2-2}
% & 1/8  \\
Method~  & mIoU  \\
\hline
N.A.~  & $74.01$  \\
CP   & $75.01$\\
DCP~  & $75.28$   \\
CSCP  & $75.63$  \\
DCSCP~  & $\bf{75.79}$   \\
\hline
\end{tabular}
\label{tab:DCSCP}
% \vspace{-0.3cm}
\end{table}

\renewcommand{\arraystretch}{1.4}
\begin{table}[]
\centering\setlength{\tabcolsep}{8pt}
\captionsetup{font={small}}
\caption{Performance analysis of different weight $\lambda$.
\vspace{-2mm}}

\footnotesize
\begin{tabular}{l|ccccc}
\hline
$\lambda$~ & $0.5$ & $0.75$ & $1$& $2$ & $4$    \\
\hline
mIoU~ & $76.20$ & $76.76$ & $76.88$& $\bf{77.60}$& $77.12$  \\
\hline
\end{tabular}
\label{tab:DCSCP}
% \vspace{-0.3cm}
\end{table}

\noindent\textbf{The trade-off weight $\lambda$.}
$\lambda$ is used for balancing the trade-off between supervised loss and unsupervised loss. The result shows that $\lambda=2$ behaves the best in our setting, where a smaller $\lambda=0.5$ will reduce a lot of useful information brought by the pseudo segmentation map. A larger $\lambda=4$ is problematic and leads to the performance decrease since the network might converge in the wrong direction.

\renewcommand{\arraystretch}{1.4}
\begin{table}[]
\centering\setlength{\tabcolsep}{32pt}
\captionsetup{font={small}}
\caption{Cross-Model network vs Cross-Head network.
\vspace{-2mm}
}
\footnotesize
\begin{tabular}{l|ccc}
\hline
Method~  & mIoU  \\
\hline
CM~  & $76.57$  \\
CH   & $\bf{77.60}$\\
\hline
\end{tabular}
\label{tab:MH}
% \vspace{-0.3cm}
\end{table}

\noindent\textbf{Cross-Head network vs Cross-Model network.}
Comparison with Cross-Head network and Cross-Model network
on CityScapes val. CH = Cross-Head network, CM=Cross-Model network. As we can see from Table \ref{tab:MH} , the Cross-Head network outperforms the Cross-Model network by +1.03$\%$. By sharing the same representation, the Cross-Head network could further improve generalization ability and then learn a more compact feature from different views.

\subsection{Qualitative Results}

In Figure \ref{fig:quantitive}, we display some qualitative results under 1/8 protocol on the Cityscapes val set, and all the approaches are based on DeepLabv3+ with ResNet-101 network. Benefited from the proposed framework with a series of components, our method shows more accurate segmentation results than the previous state-of-art method \cite{chen2021semi}, especially for the large region with complex texture and needing long-range information.

\section{Conclusion}

In this paper, we propose a novel framework called UCC (Uncertainty guided Cross-head Co-training) in the semi-supervised semantic segmentation field. Our method is the first to merge weak and strong augmentations into the cross-head co-training framework, which naturally combines the benefits of consistency and self-training. On the one hand, our proposed DCSCP boosts the consistency training samples' diversity while simultaneously tackling the biased distribution caused by the imbalanced datasets, and the gap between the labeled and unlabeled data. On the other hand, our proposed UGRM enhances the self-training pseudo labels by suppressing the effect of the low-quality pseudo labels from its peer via modeling uncertainty. We prove the effectiveness of our paradigm in semi-supervised semantic segmentation, with the spotlight on two commonly used benchmarks, including CityScapes and PASCAL VOC 2012.

Consistency regularization based methods have been well developed in the last decades, yet the effectiveness of self-training is always ignored. So how to further exploit the potential benefit of the noisy pseudo label in self-training while utilizing consistency regularization is meaningful in the future.
\newpage
%%%%%%%%% REFERENCES
{\small
\bibliographystyle{ieee_fullname}
\bibliography{seg}
}

\end{document}